\documentclass[runningheads]{paper-2330}
\usepackage{graphicx}
\usepackage{booktabs}
\usepackage{multirow}
\usepackage{amssymb}
\usepackage{amsmath}
\usepackage{enumitem}
\usepackage{xcolor}
\usepackage{comment}
\usepackage{bm}
\usepackage[nodisplayskipstretch]{setspace}
\usepackage{caption}
\usepackage{subcaption}
\usepackage{bm}
\usepackage{hyperref}
\usepackage{fontawesome}

\usepackage{calc}
\newlength{\heightofx}
\newcommand{\plus}{\settoheight{\heightofx}{x}%
\raisebox{0.5\heightofx-(0.5\totalheight-\depth)}%
{\scalebox{0.6}{+}}} 

\begin{document}

\title{How Does Pruning Impact Long-Tailed Multi-Label Medical Image Classifiers?}
\titlerunning{How Does Pruning Impact Medical Image Classifiers?}

\author{Gregory Holste\inst{1} \and
Ziyu Jiang\inst{2} \and
Ajay Jaiswal\inst{1} \and
Maria Hanna\inst{3} \and
Shlomo Minkowitz\inst{3} \and
Alan C. Legasto\inst{3} \and
Joanna G. Escalon\inst{3} \and
Sharon Steinberger\inst{3} \and
Mark Bittman \inst{3} \and
Thomas C. Shen \inst{4} \and
Ying Ding\inst{1} \and
Ronald M. Summers \inst{4} \and
George Shih\inst{3} \and
Yifan Peng\inst{3}\textsuperscript{(\faEnvelopeO)} \and
Zhangyang Wang \inst{1}\textsuperscript{(\faEnvelopeO)}}
\authorrunning{G. Holste \textit{et al.}}


\institute{The University of Texas at Austin, Austin, TX, USA\\
\email{atlaswang@utexas.edu}
\and 
Texas A\&M University, College Station, TX, USA
\and
Weill Cornell Medicine, New York, NY, USA\\
\email{yip4002@med.cornell.edu}
\and
Clinical Center, National Institutes of Health, Bethesda, MD, USA
}

\maketitle              

\begin{abstract}
Pruning has emerged as a powerful technique for compressing deep neural networks, reducing memory usage and inference time without significantly affecting overall performance. However, the nuanced ways in which pruning impacts model behavior are not well understood, particularly for \textit{long-tailed}, \textit{multi-label} datasets commonly found in clinical settings. This knowledge gap could have dangerous implications when deploying a pruned model for diagnosis, where unexpected model behavior could impact patient well-being. To fill this gap, we perform the first analysis of pruning's effect on neural networks trained to diagnose thorax diseases from chest X-rays (CXRs). On two large CXR datasets, we examine which diseases are most affected by pruning and characterize class ``forgettability" based on disease frequency and co-occurrence behavior. Further, we identify individual CXRs where uncompressed and heavily pruned models disagree, known as pruning-identified exemplars (PIEs), and conduct a human reader study to evaluate their unifying qualities. We find that radiologists perceive PIEs as having more label noise, lower image quality, and higher diagnosis difficulty. This work represents a first step toward understanding the impact of pruning on model behavior in deep long-tailed, multi-label medical image classification. All code, model weights, and data access instructions can be found at \url{https://github.com/VITA-Group/PruneCXR}.

\keywords{Pruning \and Chest X-Ray \and Imbalance \and Long-Tailed Learning}
\end{abstract}

\section{Introduction}

Deep learning has enabled significant progress in image-based computer-aided diagnosis~\cite{rajpurkar2017chexnet,hesamian2019deep,zhou2021review,Lin2022-jo,Han2022-xa}.
However, the increasing memory requirements of deep neural networks limit their practical deployment in hardware-constrained environments. 
One promising approach to reducing memory usage and inference latency is model \textbf{\textit{pruning}}, which aims to remove redundant or unimportant model weights \cite{lecun1989optimal}.
Since modern deep neural networks are often overparameterized, they can be heavily pruned with minimal impact on overall performance \cite{zhu2017prune,kurtic2022gmp,lee2018snip,frankle2018lottery}.
This being said, the impact of pruning on model behavior \textit{beyond} high-level performance metrics like top-1 accuracy remain unclear.
This gap in understanding has major implications for real-world deployment of neural networks for high-risk tasks like disease diagnosis, where pruning may cause unexpected consequences that could potentially threaten patient well-being.

To bridge this gap, this study aims to answer the following guiding questions by conducting experiments to dissect the differential impact of pruning:
\begin{enumerate}[label=Q\arabic*., topsep=1pt, leftmargin=1cm]\bfseries
    \item What is the impact of pruning on overall performance in\\long-tailed multi-label medical image classification?
    \item Which disease classes are most affected by pruning and why?
    \item How does disease co-occurrence influence the impact of pruning?
    \item Which individual images are most vulnerable to pruning?
\end{enumerate}
We focus our experiments on thorax disease classification on chest X-rays (CXRs), a challenging \textbf{\textit{long-tailed}} and \textbf{\textit{multi-label}} computer-aided diagnosis problem, where patients may present with multiple abnormal findings in one exam and most findings are rare relative to the few most common diseases \cite{holste2022long}.

This study draws inspiration from Hooker \textit{et al.} \cite{hooker2019compressed}, who found that pruning disparately impacts a small subset of classes in order to maintain overall performance. The authors also introduced \textbf{\textit{pruning-identified exemplars (PIEs)}}, images where an uncompressed and heavily pruned model disagree. They discovered that PIEs share common characteristics such as multiple salient objects and noisy, fine-grained labels.
While these findings uncover what neural networks ``forget" upon pruning, the insights are limited to highly curated natural image datasets where each image belongs to one class.
Previous studies have shown that pruning can enhance fairness \cite{wu2022fairprune}, robustness \cite{chen2021enhancing}, and efficiency for medical image classification \cite{yang2019network,fernandes2020automatic,hajabdollahi2019hierarchical} and segmentation \cite{mahbod2022deep,lin2022lighter,valverde2022sauron,jeong2021neural,dinsdale2022stamp} tasks. However, these efforts also either focused solely on high-level performance or did not consider settings with severe class imbalance or co-occurrence.

Unlike existing work, we explicitly connect class ``forgettability" to the unique aspects of our problem setting: disease frequency (long-tailedness) and disease co-occurrence (multi-label behavior).
Since many diagnostic exams, like CXR, are long-tailed and multi-label, this work fills a critical knowledge gap enabling more informed deployment of pruned disease classifiers.
We hope that our findings can provide a foundation for future research on pruning in clinically realistic settings.

\section{Methods}

\subsection{Preliminaries}

\textbf{Datasets.} For this study, we use expanded versions of NIH ChestXRay14 \cite{wangchest2017} and MIMIC-CXR \cite{johnson2019mimic}, two large-scale CXR datasets for multi-label disease classification.\footnote[1]{NIH ChestXRay14 can be found \href{https://nihcc.app.box.com/v/ChestXray-NIHCC}{here}, and MIMIC-CXR can be found \href{https://physionet.org/content/mimic-cxr/2.0.0/}{here}.}
As described in Holste \textit{et al.} \cite{holste2022long}, we augmented the set of possible labels for each image by adding five new rare disease findings parsed from radiology reports. This creates a challenging long-tailed classification problem, with training class prevalence ranging from under 100 to over 70,000 (Supplement).
\textbf{\textit{NIH-CXR-LT}} contains 112,120 CXRs, each labeled with at least one of 20 classes, while \textbf{\textit{MIMIC-CXR-LT}} contains 257,018 frontal CXRs labeled with at least one of 19 classes.
Each dataset was split into training (70\%), validation (10\%), and test (20\%) sets at the patient level.

\textbf{Model Pruning \& Evaluation.} Following Hooker \textit{et al.} \cite{hooker2019compressed}, we focus on global unstructured \textbf{\textit{L1 pruning}} \cite{zhu2017prune}. After training a disease classifier, a fraction $k$ of weights with the smallest magnitude are ``pruned" (set to zero); for instance, $k=0.9$ means 90\% of weights have been pruned. While area under the receiver operating characteristic curve is a standard metric on related datasets \cite{rajpurkar2017chexnet,seyyed2020chexclusion,wangchest2017}, it can become heavily inflated in the presence of class imbalance \cite{fernandez2018learning,davis2006relationship}. Since we seek a metric that is both resistant to imbalance and captures performance across thresholds (as choosing a threshold is non-trivial in the multi-label setting \cite{rethmeier2022long}), we use \textbf{average precision (AP)} as our primary metric.

\subsection{Assessing the Impact of Pruning}

\textbf{Experimental Setup.} We first train a baseline model to classify thorax diseases on both NIH-CXR-LT and MIMIC-CXR-LT. The architecture used was a ResNet50 \cite{he2016deep} with ImageNet-pretrained weights and a sigmoid cross-entropy loss. For full training details, please see the Supplemental Materials and code repository. Following Hooker \textit{et al.} \cite{hooker2019compressed}, we then repeat this process with 30 unique random initializations, performing L1 pruning at a range of sparsity ratios $\bm{k \in \{0, 0.05,} \textbf{\dots, }\bm{0.9, 0.95\}}$ on each model and dataset. Using a ``population" of 30 models allows for reliable estimation of model performance at each sparsity ratio. We then analyze how pruning impacts overall, disease-level, and image-level model behavior with increasing sparsity as described below.

\textbf{Overall \& Class-Level Analysis.} To evaluate the overall impact of pruning, we compute the mean AP across classes for each sparsity ratio and dataset. We use Welch's t-test to assess performance differences between the 30 uncompressed models and 30 $k$-sparse models.
We then characterize the class-level impact of pruning by considering the relative change in AP from an uncompressed model to its $k$-sparse counterpart for all $k$. Using relative change in AP allows for comparison of the \textit{impact} of pruning regardless of class difficulty. We then define the \textbf{\textit{forgettability curve}} of a class $c$ as follows:
\begin{equation}
    \left\{\textrm{med}\left\{ \frac{\textrm{AP}_{i,k,c} - \textrm{AP}_{i,0,c}}{\textrm{AP}_{i,0,c}} \right\}_{i \in \{1, \dots, 30\}} \right\}_{k \in \{0, 0.05, \dots, 0.9, 0.95\}}
\end{equation}
where $\textrm{AP}_{i,k,c} :=$ AP of the $i$\textsuperscript{th} model with sparsity $k$ on class $c$, and $\textrm{med}(\cdot) :=$ median across all 30 runs. We analyze how these curves relate to class frequency and co-occurrence using Pearson ($r$) and Spearman ($\rho$) correlation tests.

\begin{figure}[!ht]
    \centering    
    \includegraphics[scale=0.48]{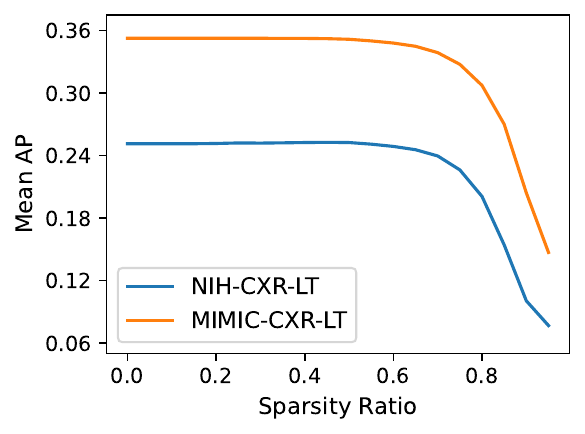}
    \includegraphics[scale=0.49]{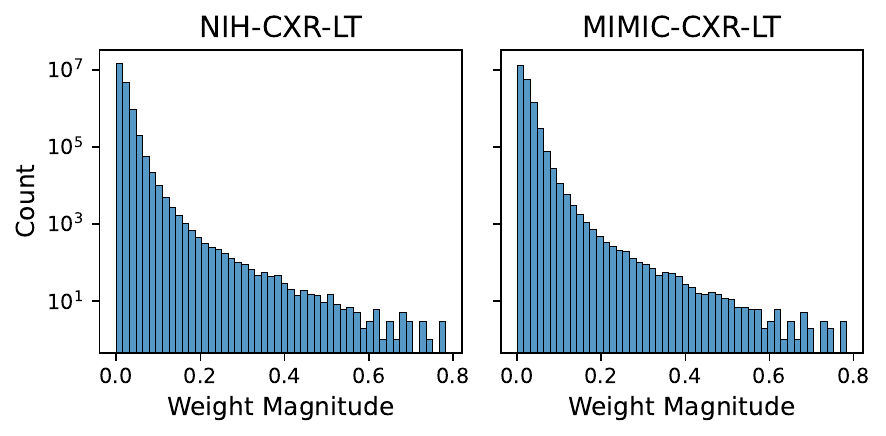}
    \caption{Overall effect of pruning on disease classification performance. Presented is the mean AP (median across 30 runs) for sparsity ratios $k \in \{0, \dots, 0.95\}$ (left) and log-scale histogram of model weight magnitudes (right).}
    \label{fig:overall_prune}
\end{figure}

\textbf{Incorporating Disease Co-occurrence Behavior.}
For each \textit{unique pair} of NIH-CXR-LT classes, we compute the \textbf{\textit{Forgettability Curve Dissimilarity (FCD)}}, the mean squared error (MSE) between the forgettability curves of each disease. FCD quantifies how similar two classes are with respect to their forgetting behavior over all sparsity ratios. Ordinary least squares (OLS) linear regression is employed to understand the interaction between difference in class frequency and class co-occurrence with respect to FCD for a given disease pair.

\subsection{Pruning-Identified Exemplars (PIEs)}

\textbf{Definition.} After evaluating the overall and class-level impact of pruning on CXR classification, we investigate which individual images are most vulnerable to pruning. Like Hooker \textit{et al.} \cite{hooker2019compressed}, we consider PIEs to be images where an uncompressed and pruned model disagree. Letting $C$ be the number of classes, we compute the average prediction $\frac{1}{30} \sum_i \bm{\hat{y}_0} \in \mathbb{R}^C$ of the uncompressed models and average prediction $\frac{1}{30} \sum_i \bm{\hat{y}_{0.9}} \in \mathbb{R}^C$ of the L1-pruned models at 90\% sparsity for all NIH-CXR-LT test set images. Then the Spearman rank correlation $\sigma(\frac{1}{30} \sum_i \bm{\hat{y}_0}, \frac{1}{30} \sum_i \bm{\hat{y}_{0.9}})$ represents the agreement between the uncompressed and heavily pruned models for each image; we define PIEs as images whose correlation falls in the bottom 5\textsuperscript{th} percentile of test images.

\textbf{Analysis \& Human Study.} To understand the common characteristics of PIEs, we compare how frequently (i) each class appears and (ii) images with $d=0,\dots,3,4+$ simultaneous diseases appear in PIEs relative to non-PIEs. To further analyze qualities of CXRs that require domain expertise, we conducted a human study to assess radiologist perceptions of PIEs. Six board-certified attending radiologists were each presented with a unique set of 40 CXRs (half PIE, half non-PIE). Each image was presented along with its ground-truth labels and the following three questions:
\begin{enumerate}[label=\textbf{\arabic*}., topsep=0pt, leftmargin=1cm]
    \item \textbf{Do you fully agree with the given label?} [Yes/No]
    \item \textbf{How would you rate the image quality?} [1-5 Likert]
    \item \textbf{How difficult is it to properly diagnose this image?} [1-5 Likert]
\end{enumerate}
We use the Kruskal-Wallis test to evaluate differential perception of PIEs.

\begin{figure}[!ht]
    \centering
    \includegraphics[scale=0.525]{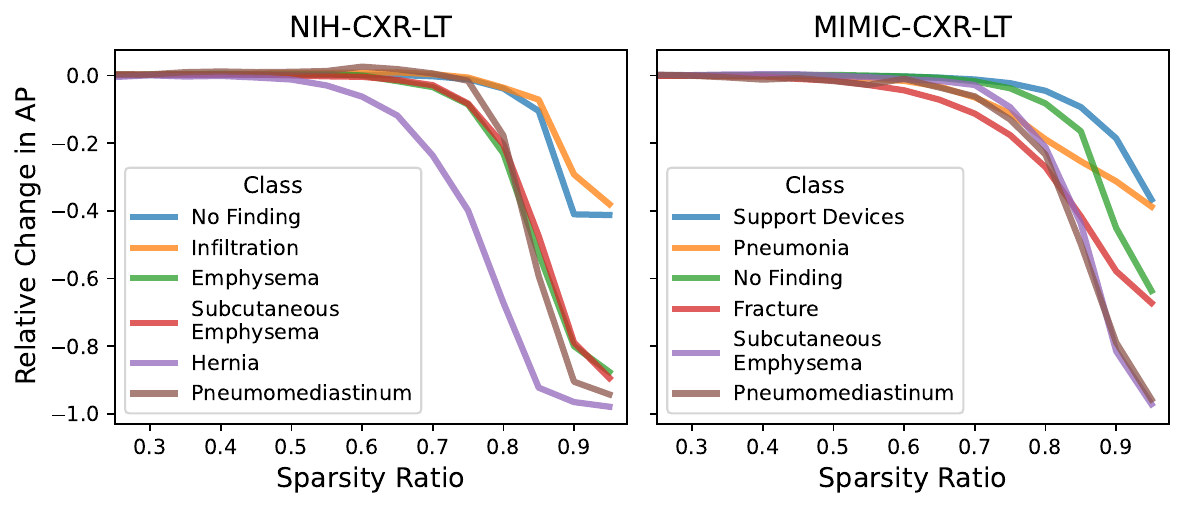}
    \caption{``Forgettability curves" depicting relative change in AP (median across 30 runs at each sparsity ratio) upon L1 pruning for a subset of classes.}
    \label{fig:forgettability_curves}
\end{figure}

\section{Results}

\subsection{What is the overall effect of pruning?}

We find that under L1 pruning, the first sparsity ratio causing a significant drop in mean AP is 65\% for NIH-CXR-LT ($P<0.001$) and 60\% for MIMIC-CXR-LT ($P<0.001$) (Fig. \ref{fig:overall_prune}, left).
This observation may be explained by the fact that ResNet50 is highly overparameterized for this task. Since only a subset of weights are required to adequately model the data, the trained classifiers have naturally sparse activations (Fig. \ref{fig:overall_prune}, right). For example, over half of all learned weights have magnitude under 0.01.
However, beyond a sparsity ratio of 60\%, we observe a steep decline in performance with increasing sparsity for both datasets.

\begin{figure}[!ht]
    \centering
    \includegraphics[scale=0.55]{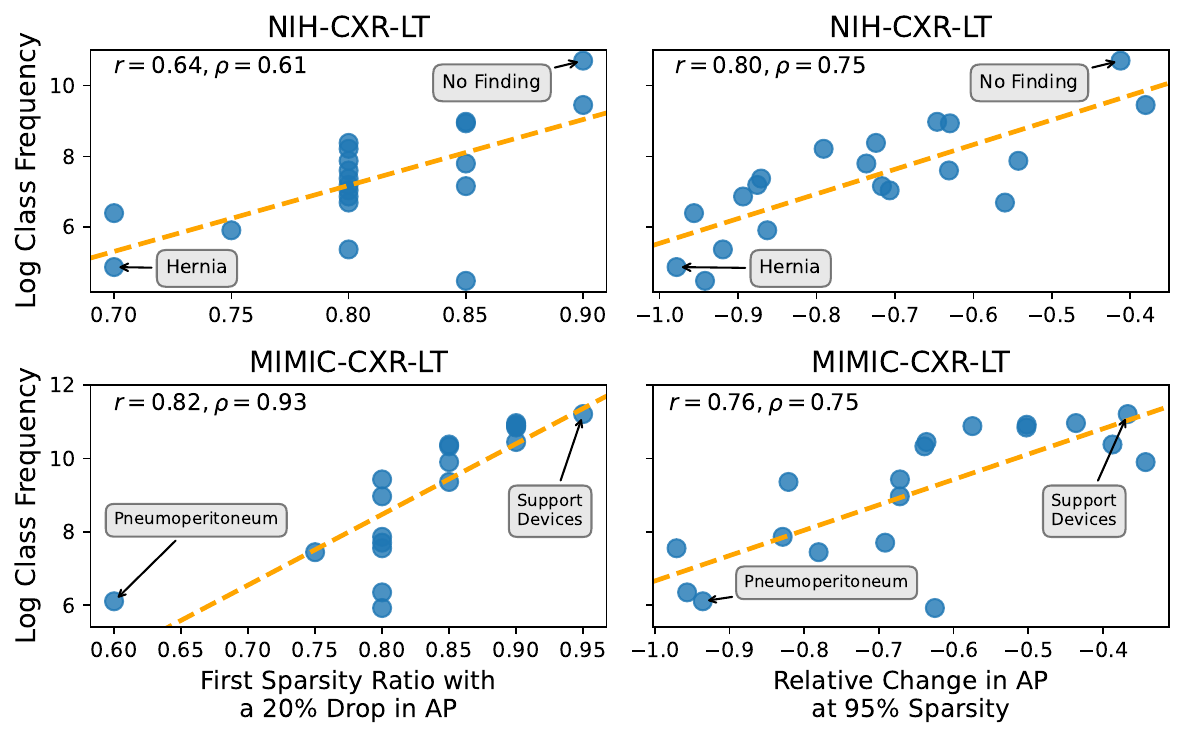}
    \caption{Relationship between class ``forgettability" and frequency. We characterize which classes are forgotten \textit{first} (left) and which are \textit{most} forgotten (right).}
    \label{fig:class_prune_corr}
\end{figure}

\begin{figure}[!ht]
    \centering
    \includegraphics[scale=0.6]{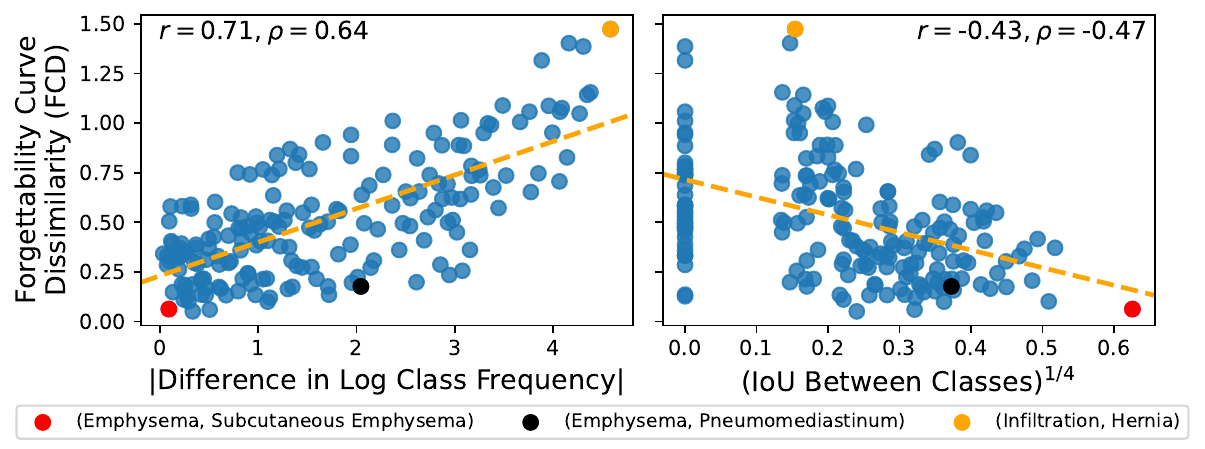}
    \caption{Mutual relationship between pairs of diseases and their forgettability curves. For each pair of NIH-CXR-LT classes, FCD is plotted against the absolute difference in log frequency (left) and the IoU between the two classes (right).}
    \label{fig:mutual_class_prune_corr}
\end{figure}

\subsection{Which diseases are most vulnerable to pruning and why?}

Class forgettability curves in Fig. \ref{fig:forgettability_curves} depict the relative change in AP by sparsity ratio for a representative subset of classes. Although these curves follow a similar general trend to Fig. \ref{fig:overall_prune}, some curves (i) drop earlier and (ii) drop more considerably  at high sparsity. Notably, we find a strong positive relationship between training class frequency and (i) the first sparsity ratio at which a class experienced a median 20\% relative drop in AP ($\rho=0.61, P=0.005$ for NIH-CXR-LT; $\rho=0.93, P \ll 0.001$ for MIMIC-CXR-LT) and (ii) the median relative change in AP at 95\% sparsity ($\rho=0.75, P<0.001$ for NIH-CXR-LT; $\rho=0.75, P<0.001$ for MIMI-CXR-LT). These findings indicate that, in general, \textbf{rare diseases are forgotten earlier} (Fig. \ref{fig:class_prune_corr}, left) \textbf{and are more severely impacted at high sparsity} (Fig. \ref{fig:class_prune_corr}, right).

\subsection{How does disease co-occurrence influence class forgettability?}

Our analysis reveals that for NIH-CXR-LT, the absolute difference in log test frequency between two diseases is a strong predictor of the pair's FCD ($\rho=0.64, P \ll 0.001$). This finding suggests that \textbf{diseases with larger differences in prevalence exhibit more distinct forgettability behavior} upon L1 pruning (Fig. \ref{fig:mutual_class_prune_corr}, left). To account for the multi-label nature of thorax disease classification, we also explore the relationship between intersection over union (IoU) -- a measure of co-occurrence between two diseases -- and FCD. Our analysis indicates that the IoU between two diseases is negatively associated with FCD ($\rho=-0.47, P \ll 0.001$). This suggests that \textbf{the more two diseases co-occur, the more similar their forgetting trajectories are across all sparsity ratios} (Fig. \ref{fig:mutual_class_prune_corr}, right). For example, the disease pair (Infiltration, Hernia) has a dramatic difference in prevalence ($|\textrm{LogFreqDiff}|=4.58$) \textit{and} rare co-occurrence ($\textrm{IoU}^{1/4}=0.15$), resulting in an extremely high FCD for the pair of diseases.

We also find, however, that there is a push and pull between differences in individual class frequency and class co-occurrence with respect to FCD. To illustrate, consider the disease pair (Emphysema, Pneumomediastinum) marked in black in Figure \ref{fig:mutual_class_prune_corr}. These classes have an absolute difference in log frequency of 2.04, which would suggest an FCD of around 0.58. However, because Emphysema and Pneumomediastinum co-occur relatively often ($\textrm{IoU}^{1/4}=0.37$), their forgettability curves are \textit{more similar than prevalence alone would dictate}, resulting in a lower FCD of 0.18. To quantify this effect, we obtain an OLS model that fitted FCD as a function of $|\textrm{LogFreqDiff}|$, $\textrm{IoU}^{1/4}$, and their interaction:
\begin{equation}
    \text{FCD} = 0.27 + 0.21 |\text{LogFreqDiff}| - 0.05 (\text{IoU})^{1/4} - 0.31 |\text{LogFreqDiff}|*(\text{IoU})^{1/4}
\end{equation}

We observe a statistically significant interaction effect between the difference in individual class frequency and class co-occurrence on FCD ($\beta_3 = -0.31, P=0.005$). Thus, for disease pairs with a very large difference in prevalence, the effect of co-occurrence on FCD is even more pronounced (Supplement).

\begin{figure}[!ht]
    \centering
    \includegraphics[scale=0.475]{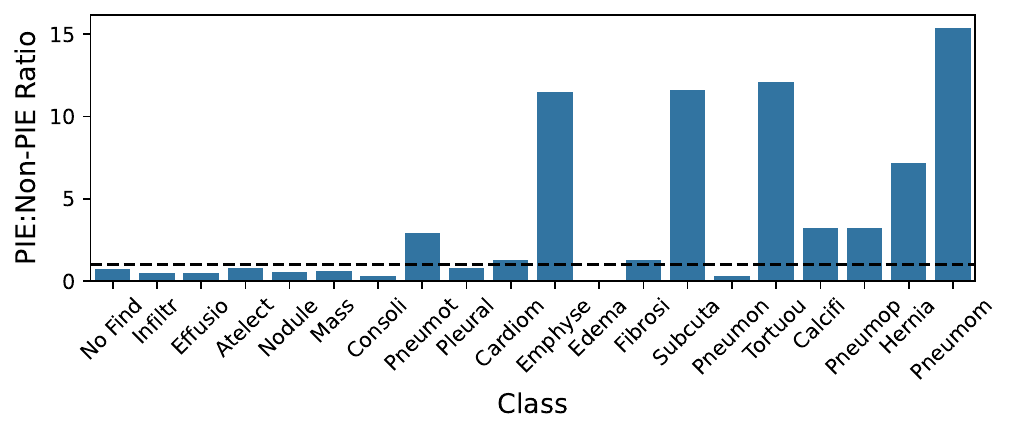}
    \includegraphics[scale=0.475]{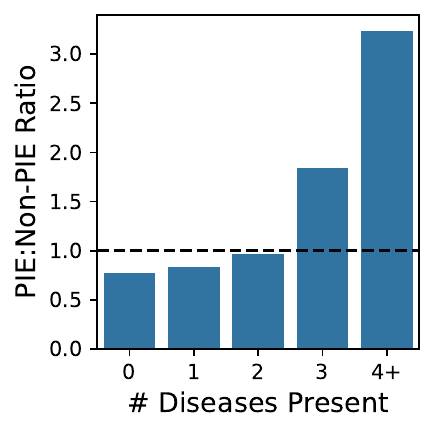}
    \caption{Unique characteristics of PIEs. Presented is the \textit{ratio} of class prevalence (left) and number of diseases per image (right) in PIEs relative to non-PIEs. The dotted line represents the 1:1 ratio (equally frequent in PIEs vs. non-PIEs).}
    \label{fig:pie_analysis}
\end{figure}

\begin{figure}[!ht]
    \centering
    \includegraphics[scale=0.48]{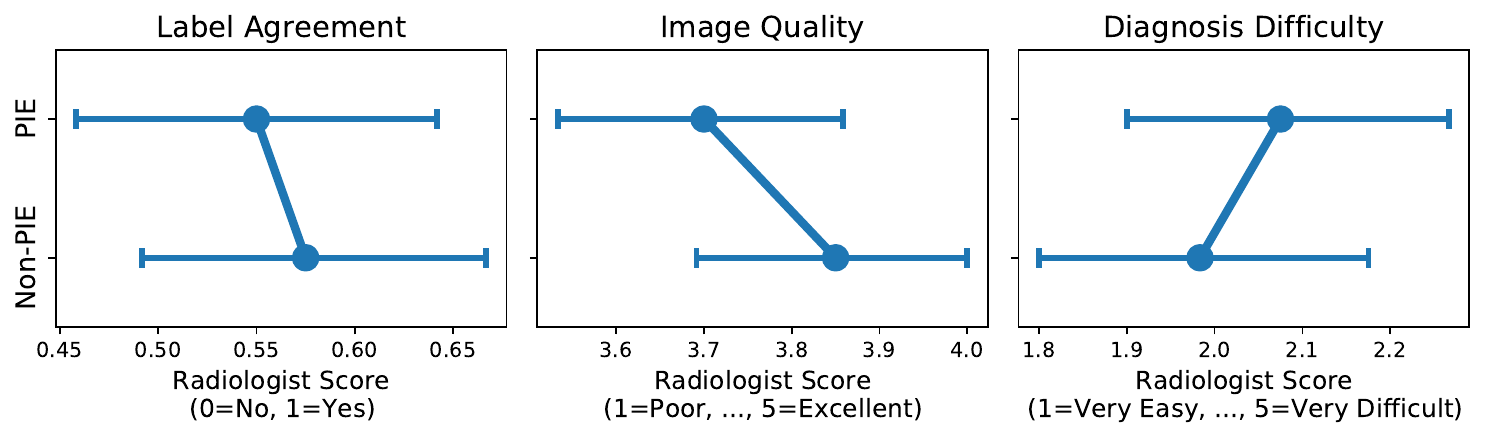}
        \caption{Human study results describing radiologist perception of PIEs vs. non-PIEs. Mean $\pm$ standard deviation (error bar) radiologist scores are presented.}
    \label{fig:pie_survey}
\end{figure}

\subsection{What do pruning-identified CXRs have in common?}
For NIH-CXR-LT, we find that \textbf{PIEs are more likely to contain rare diseases} and \textbf{more likely to contain 3\plus \ simultaneous diseases} when compared to non-PIEs (Fig. \ref{fig:pie_analysis}). The five rarest classes appear 3-15x more often in PIEs than non-PIEs, and images with 4\textbf{\plus} diseases appear 3.2x more often in PIEs.

In a human reader study involving $240$ CXRs from the NIH-CXR-LT test set (120 PIEs and 120 non-PIEs), \textbf{radiologists perceived that PIEs had more label noise, lower image quality, and higher diagnosis difficulty} (Fig. \ref{fig:pie_survey}). However, due to small sample size and large variability, these differences are not statistically significant. Respondents fully agreed with the label 55\% of the time for PIEs and 57.5\% of the time for non-PIEs ($P=0.35$), gave an average image quality of 3.6 for PIEs and 3.8 for non-PIEs ($P=0.09$), and gave an average diagnosis difficulty of 2.5 for PIEs and 2.05 for non-PIEs ($P=0.25$).

Overall, these findings suggest that pruning identifies CXRs with many potential sources of difficulty, such as containing underrepresented diseases, (partially) incorrect labels, low image quality, and complex disease presentation.

\section{Discussion \& Conclusion}
In conclusion, we conducted the first study of the effect of pruning on multi-label, long-tailed medical image classification, focusing on thorax disease diagnosis in CXRs. Our findings are summarized as follows:
\begin{enumerate}
    \item As observed in standard image classification, CXR classifiers can be heavily pruned (up to 60\% sparsity) before dropping in overall performance.
    \item Class frequency is a strong predictor of both \textit{when} and \textit{how severely} a class is impacted by pruning. Rare classes suffer the most.
    \item Large differences in class frequency lead to dissimilar ``forgettability" behavior and stronger co-occurrence leads to more similar forgettability behavior.
    \begin{itemize}[topsep=0pt]
        \item Further, we discover a significant interaction effect between these two factors with respect to how similarly pruning impacts two classes.
    \end{itemize}
    \item We adapt PIEs to the multi-label setting, observing that PIEs are far more likely to contain rare diseases and multiple concurrent diseases.
    \begin{itemize}[topsep=0pt]
        \item A radiologist study further suggests that PIEs have more label noise, lower image quality, and higher diagnosis difficulty.
    \end{itemize}
\end{enumerate}

It should be noted that this study is limited to the analysis of global unstructured L1 (magnitude-based) pruning, a simple heuristic for post-training network pruning. Meanwhile, other state-of-the-art pruning approaches \cite{kurtic2022gmp,lee2018snip,frankle2018lottery} and model compression techniques beyond pruning (e.g., weight quantization \cite{jacob2018quantization} and knowledge distillation \cite{hinton2015distilling}) could be employed to strengthen this work. Additionally, since our experiments only consider the ResNet50 architecture, it remains unclear whether other training approaches, architectures, or compression methods could mitigate the adverse effects of pruning on rare classes. In line with recent work~\cite{jaiswal2023attend,jiang2021self,kong2023pruning}, future research may leverage the insights gained from this study to develop an algorithm for improved long-tailed learning on medical image analysis tasks. For example, PIEs could be interpreted as salient, difficult examples that warrant greater weight during training. Conversely, PIEs may just as well be regarded as noisy examples to be ignored, using pruning as a tool for data cleaning.

\section*{Acknowledgments}

This project was supported by the Intramural Research Programs of the National Institutes of Health, Clinical Center. It also was supported by the National Library of Medicine under Award No. 4R00LM013001, NSF CAREER Award No. 2145640, Cornell Multi-Investigator Seed Grant (Peng and Shih), and Amazon Research Award. 

%
%
%
\bibliographystyle{paper-2330}
\bibliography{paper-2330}

\newpage

\section*{Supplementary Materials}

\begin{table}[!ht]
	\centering
	\caption{Summary of implementation details for training. LR = learning rate.}
	\begin{tabular}{l|c}
		\toprule
		\textbf{Model} & ResNet50 (ImageNet-pretrained) \\
		\textbf{Optimizer (LR)} & Adam ($1 \times 10^{-4}$) \\
		\textbf{Augmentations} & Random horizontal flip, Random rotation (-15$^\circ$, 15$^\circ$) \\
		\textbf{Preprocessing} & ImageNet normalization, Resize to $256 \times 256$ \\
		\textbf{Early Stopping} & 15 epochs with no improvement in validation AUC \\
		\textbf{Framework} & PyTorch \\
		\textbf{Hardware} & 1 NVIDIA RTX A6000 GPU \\
		\textbf{Training Walltime} & $\sim$3 hours (NIH-CXR-LT), $\sim$6.5 hours (MIMIC-CXR-LT) \\
		\bottomrule
	\end{tabular}
\end{table}


\begin{figure}[!ht]
	\centering
	\includegraphics[scale=0.5]{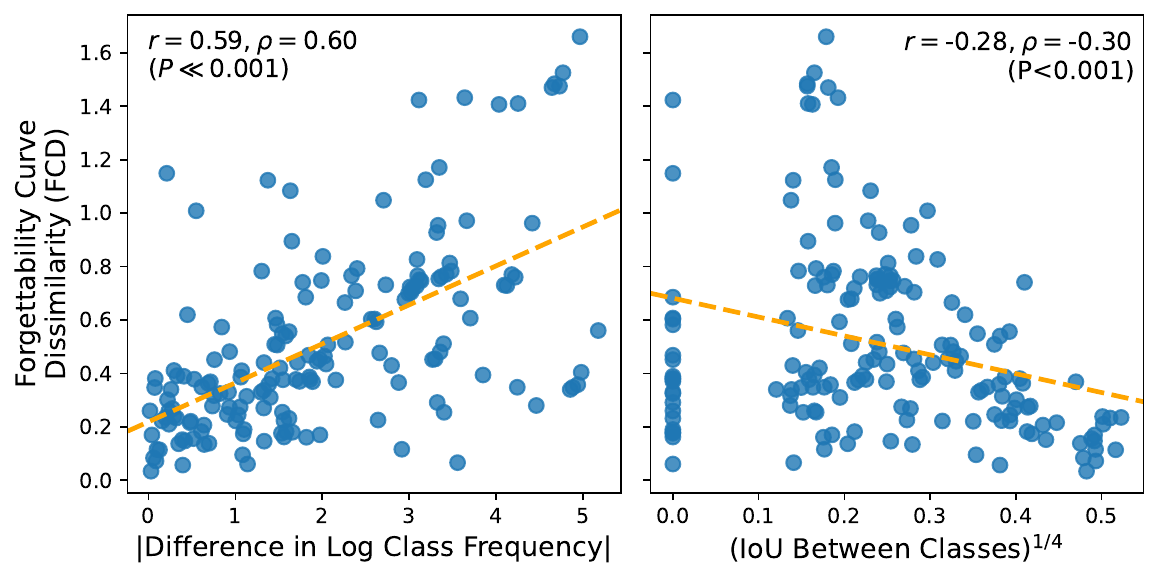}
	\caption{Mutual relationship between pairs of diseases and their forgettability curves. For each pair of MIMIC-CXR-LT classes, FCD is plotted against the absolute difference in log frequency (left) and IoU between the two classes (right). Pearson ($r$) and Spearman ($\rho$) correlation coefficients shown, with P-value for $\rho$.}
\end{figure}

\begin{table}[ht!]
	\centering
	\renewcommand{\arraystretch}{1.1}
	\caption{Long-tailed distribution of NIH-CXR-LT and MIMIC-CXR-LT labels by data split. Italicized labels with an asterisk before the name denote the five newly added rare disease findings.}
	\begin{tabular}{lrrrr|rrrr}
		\toprule
		\multicolumn{1}{c}{Label} & \multicolumn{4}{c}{NIH-CXR-LT} & \multicolumn{4}{c}{MIMIC-CXR-LT} \\
		\cmidrule(lr){2-5} \cmidrule(lr){6-9}
		& \multicolumn{1}{c}{Train} & \multicolumn{1}{c}{Val} & \multicolumn{1}{c}{Test} & \multicolumn{1}{c}{\textbf{Total}} & \multicolumn{1}{|c}{Train} & \multicolumn{1}{c}{Val} & \multicolumn{1}{c}{Test} & \multicolumn{1}{c}{\textbf{Total}} \\
		\midrule
		No Finding & 44625 & 6766 & 8015 & 59406 & 34380 & 3768 & 9828 & 47976 \\
		Support Devices & - & - & - & - & 73641 & 8328 & 22397 & 104366 \\
		Lung Opacity & - & - & - & - & 57591 & 6397 & 17405 & 81393 \\
		Infiltration & 12739 & 1996 & 5159 & 19894 & - & - & - & - \\
		Atelectasis & 7587 & 1272 & 2700 & 11559 & 51055 & 5595 & 15102 & 71752 \\
		Effusion & 7919 & 1663 & 3735 & 13317 & 53029 & 5822 & 16300 & 75151 \\
		Nodule & 4359 &  667 & 1305 &  6331 & - & - & - & - \\
		Mass & 3689 &  764 & 1329 &  5782 & - & - & - & - \\
		Pneumothorax & 2432 &  764  & 2106  & 5302 & 11579 & 1388  & 3517  & 16484 \\
		Consolidation & 2626 &  544 & 1497 &  4667 & 12406 & 1413 &  3776 &  17595 \\
		Cardiomegaly & 1590  & 318  & 868  & 2776 & 55020 & 6026 & 16747  & 77793 \\
		Pleural Thickening & 1998 &  485  & 902  & 3385 & - & - & - & - \\
		Fibrosis & 1138  & 183  & 365 &  1686 & - & - & - & - \\
		Edema & 1283 &  269 &  751 &  2303 & 30555 & 3502  & 9530  & 43587 \\
		Emphysema & 1327 &  272  & 917  & 2516 & - & - & - & - \\
		Pneumonia & 806  & 173  & 452 &  1431 & 32131  & 3589 &  9455  & 45175 \\
		$^\star$\textit{Subcutaneous} & 957 &  221 &  813 &  1991 & 1900 &  269  &  599  &  2768 \\
		\ \ \textit{Emphysema} &  &  &  &  &  &  &  &  \\
		Enlarged Cardio & - & - & - & - & 19880 & 2188 &  6173 &  28241 \\
		\ \ -mediastinum &  &  &  &  &  &  &  &  \\
		Fracture & - & - & - & - & 7823 &  862 &  2410 &  11095 \\
		Lung Lesion & - & - & - & - & 1707 &  172  &  526   & 2405 \\
		$^\star$\textit{Tortuous Aorta} & 598  &  49  &  95  &  742 & 2212 &  233  &  641  &  3086 \\
		$^\star$\textit{Calcification of} & 368  &  32  &  55  &  455 & 2595 &  282  &  811  &  3688 \\
		\ \ \ \textit{the Aorta} &  &  &  &  &  &  &  &  \\
		$^\star$\textit{Pneumo-} & 214  &  33  &  69  &  316 & 448  &  59  &  134  &   641 \\
		\ \ \  \textit{peritoneum} & & & & & & & & \\
		$^\star$\textit{Pneumo-} & 88  &  22 &  143  &  253 & 571  &  87  &  195  &   853 \\
		\ \ \ \textit{mediastinum} & & & & & & & & \\
		Hernia & 130  &  35  &  62  &  227 & - & - & - & - \\
		Pleural Other & - & - & - & - & 373  &  42  &  138  &   553 \\
		\midrule \textbf{Total} & 78506 & 12533 & 21081 & 112120 & 182386 & 20363 & 54269 & 257018 \\
		\bottomrule
	\end{tabular}
\end{table}

\end{document}